\documentclass[a4paper]{article}
\usepackage{float}
\usepackage[pages=all, color=black, position={current page.south}, placement=bottom, scale=1, opacity=1, vshift=5mm]{background}
\SetBgContents{
	
}

\usepackage{enumitem}
\setitemize{noitemsep,topsep=2pt,parsep=2pt,partopsep=2pt}

\usepackage{subcaption}

\usepackage[margin=1in]{geometry} 

\usepackage{amsmath}
\usepackage{amsthm}
\usepackage{amssymb}

\usepackage{biblatex}
\usepackage[utf8]{inputenc}
\usepackage{hyperref}
\hypersetup{
	unicode,
	pdfauthor={Jacob Zietek, Nicholas Wade, Cole Roberts, Aref Malek, Manish Pylla, Will Xu, Sagar Patil},
	pdftitle={Pac-Man Pete: An extensible framework for building AI in VEX Robotics},
	pdfsubject={robotics},
	pdfkeywords={robotics, AI, educational robotics, sim2real, computer vision, VEX Robotics, open source},
	pdfproducer={LaTeX},
	pdfcreator={pdflatex}
}

\theoremstyle{plain}

\theoremstyle{definition}

\usepackage{graphicx, color}
\graphicspath{{fig/}}

\usepackage{algorithm, algpseudocode} 
\usepackage{mathrsfs} 

\usepackage{lipsum}

\title{Pac-Man Pete: An extensible framework for building AI in VEX Robotics}
\author{Jacob Zietek$^{1,1}$ \and Nicholas Wade$^{1,2}$ \and Cole Roberts$^{2,3}$ \and Sagar Patil$^{1,4}$ \and Aref Malek$^{1,5}$ \and Manish Pylla$^{1,6}$ \and Will Xu$^{2,7}$}

\date{
	$^1$Purdue University, Department of CS \\
	$^2$Purdue University, Department of ECE\\ \texttt{\{jzietek$^1$, nwade$^2$, rober638$^3$, patilsr$^4$ maleka$^5$, mpylla$^6$, xu1321$^7$\}@purdue.edu}\\%
}

\begin{document}
	\maketitle
	
	\begin{abstract}

        This technical report details VEX Robotics team BLRSAI’s development of a fully autonomous robot for VEX Robotics’ Tipping Point AI Competition. We identify and develop three separate critical components. This includes a Unity simulation and reinforcement learning model training pipeline, a malleable computer vision pipeline, and a data transfer pipeline to offload large computations from the VEX V5 Brain/micro-controller to an external computer. We give the community access to all of these components in hopes they can reuse and improve upon them in the future, and that it’ll spark new ideas for autonomy as well as the necessary infrastructure and programs for AI in educational robotics.
\newline
        
	\end{abstract}

    \newpage
	\tableofcontents
	\newpage
	\section{Introduction}
	\label{sec:intro}
	
	This technical report details VEX Robotics team BLRSAI’s development of a fully autonomous robot for VEX Robotics’ Tipping Point AI Competition (VAIC Tipping Point). Training robots to complete complex tasks with varied details and environments is a difficult and long process. Unlike stationary robots that can complete their designated tasks with a defined set of instructions, these robots need to make decisions based on information gathered around them using sensors and communicating with other robots. Training these robots to make correct decisions with the information they collect takes millions of iterations with feedback. These robots must also have an adequate curiosity to explore a variety of new game states. Gathering and evaluating this data manually would not only be daunting and tedious, but virtually impossible. This led us to develop a simulation to train a robot virtually. This also leaves room for developing reusable components that are all useful on their own to use this AI on VEX Robotics’ platforms. For the competition our parent team, BLRS2, provided our robot.  
	
    Due to the complex nature of the Tipping Point game, with its multiple objectives and ways to score points, we decided to simplify the goals for our AI. Instead of playing the entire Tipping Point game, our goal was to have our AI collect purple rings off the game floor and score them on pre-loaded mobile goals. These goals were pre-loaded during the 15 second “isolation” period, where the robots can not be in physical contact with the robots of the other team. This meant a pre-planned route could pick up the mobile goal. The system would need to detect the rings and drive over them in the best order. The path planning portion of this problem is already solved-- you can plan a path to collect all detected rings using simple motion planning algorithms. Since this is a relatively simple behavior to learn, this left us with a realistic goal to reach while we developed the more technically challenging parts of our system. 
    
    Due to the nature of VAIC Tipping Point the robot has to be constantly making decisions (continuous actions), which pairs well with reinforcement learning. Since the model is not making binary or probability decisions, reinforcement learning is great at making a large number of decisions trend in a certain way i.e., finding and moving towards purple rings. This also allows the robot to explore behaviors not explicitly defined, like defensively pushing enemy robots to prevent them from scoring rings. 
    
    We identify and develop three separate critical components to make this system work, detailed in Figure 1: simulation and reinforcement learning model training \cite{VEXAI-Sim}, object tracking \cite{rings_localization_mapping}, and a data transfer pipeline to offload large computations from the VEX V5 Brain/micro-controller \cite{competition_code}.
    
    \begin{figure}[H]
		\centering
		\includegraphics[width=.8\textwidth]{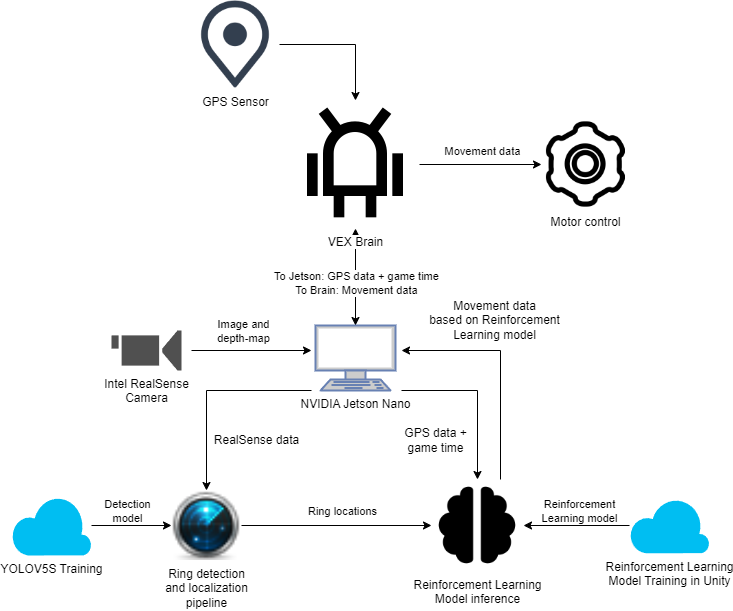}
		\caption{System diagram}
		\label{fig:system_diagram}
	\end{figure}
	
    We used the Unity Game Engine and its machine-learning agents \cite{https://doi.org/10.48550/arxiv.1809.02627} toolkit to create a simulation of the Tipping Point game. This allowed us to set up a simulation and train our robot on the same platform. We trained a custom YOLOv5s model to detect rings, and we used an Intel RealSense for the image and depth map needed to localize the rings. These rings’ relative positions were calculated to be fed to the trained reinforcement learning model. We developed a data transfer protocol using Purdue’s PROS \cite{PROS} (PROS Robotics Operating System) between the VEX Robotics brain and a NVIDIA Jetson Nano, where the computer vision pipeline and main reinforcement learning model are run. The methods and the efficacy of these components are detailed in the following sections.
	
    We give the community access to all of these components in hopes they can reuse and improve upon them in the future, and that it’ll spark new ideas for autonomy as well as the necessary infrastructure and programs for AI in educational robotics. 
	
	\section{Simulation and Training Methodology}
	\label{sec:simulation}

	The simulation team was focused on creating a realistic simulation of the Tipping Point game and training a digital twin-- which would then control the movement of our real robot \cite{VEXAI-Sim}. The simulation was built from 3D models of the field objects given by VEX along with simple cube representations of the robots, sized according to competition rules. We used Unity's machine-learning package \cite{https://doi.org/10.48550/arxiv.1809.02627} to train a reinforcement learning model in Unity's native C\# language. The agents were the robots in the simulation and rewards from Figure \ref{fig:reward_fcn} were given from the competition rules. A video of the agents training can be found \underline{\href{https://www.youtube.com/watch/-JWLU--5bbQ?feature=share}{at this link}}. Our action space was a forward velocity and a angular velocity-- for forward/backward movement and turning. The observation space is explained further in Sections 2.1 and 2.3. Training a reinforcement learning model in a virtual environment allowed the team to quickly and efficiently train and test our models.
	
	
	\subsection{Simulation}
	    We were able to create a realistic training environment for the agents using Unity’s physics engine. The 3D models that were used in the creation of the virtual playing field were taken directly from VEX and allowed us to create a true to life playing field with each object's true size and weights. After collecting all the 3D models, we put them into the game space and arranged them according to how the playing field would be set up in the competition.

	    \begin{figure}[H]
	        \centering
            \begin{subfigure}[b]{1\textwidth}
                \centering
                \includegraphics[width=.5\textwidth]{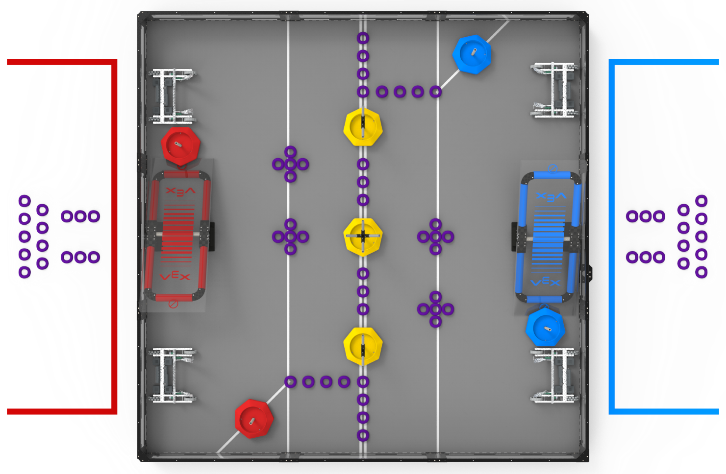}
                \caption{VEX Game Manual Image}
                \label{fig:y equals x}
            \end{subfigure}
            \newline
            \newline
            \centering
            \begin{subfigure}[b]{0.45\textwidth}
                \centering
                \includegraphics[width=1\textwidth]{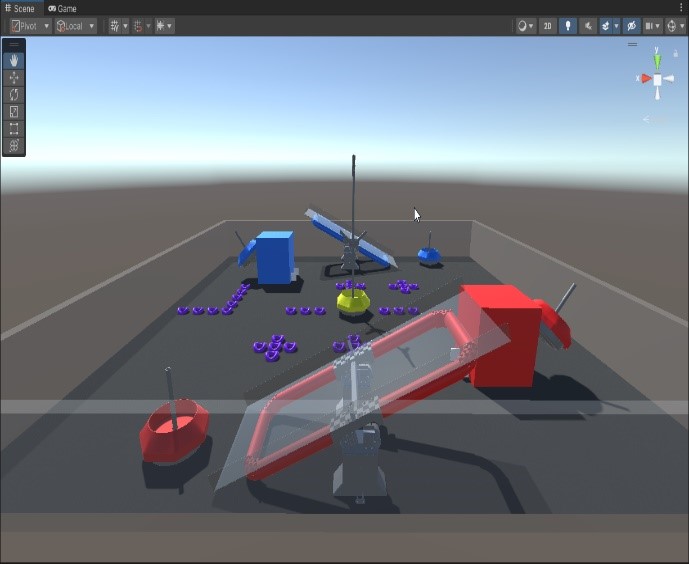}
                \caption{Unity Simulation Side View}
                \label{fig:y equals x}
            \end{subfigure}
            \hfill
            \begin{subfigure}[b]{0.45\textwidth}
                \centering
                \includegraphics[width=.83\textwidth]{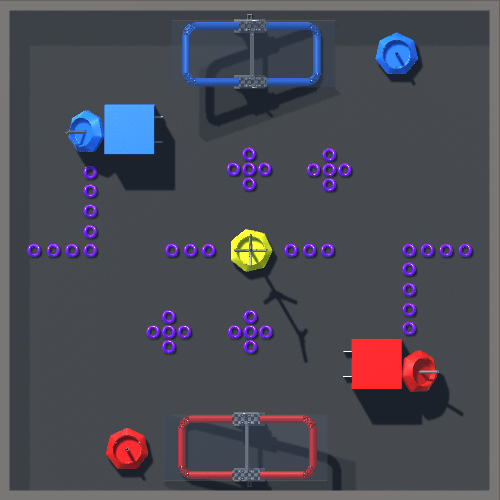}
                \caption{Unity Simulation Top-down View}
                \label{fig:y equals x}
            \end{subfigure}
	        \label{fig:my_label}
	        \caption{Playing field images}
	    \end{figure}
	
	Using a simulated environment allows you to replicate real-world inputs virtually: game time, rings in view of the robot and their respective distances, game score, and location of the robot. The game time, score, ring distance, and the robot's position are trivial using Unity’s engine. The rings in view of the robot were determined by frustum culling and occlusion culling, to simulate the limitations of the detection algorithms in Section 3. The robots had basic Unity object movement code and box colliders.
	
	\subsection{Training The Reinforcement Learning Model}
	
	The model we used was implemented by Unity’s ML-Agents toolkit \cite{https://doi.org/10.48550/arxiv.1809.02627}. Unity’s API provides access to Soft Actor Critic (SAC) and Proximal Policy Optimization (PPO) models. We made an arbitrary choice to use the PPO model. 

    Training was largely handled by Unity’s ML-Agents API. We performed some trial and error with hyperparameters, which was largely a compromise between training and evaluation time, and stability of the model during training. During training, we configured ML-Agents to run the simulation faster than real-time. Combined with the small model size, this allowed us to quickly iterate over hyperparameters.
    
    We train our model in a self-play one-versus-one scenario, as shown in Figure 2. Hyperparameters are available in the VEXAI-Sim repository \cite{VEXAI-Sim} in \textbf{mlagents-config.yaml}. Each game takes 105 seconds or 1260 steps to complete. We train for 699,300 steps or 555 games. This equates to about 16.19 hours of real-world game time.
    


    \subsection{Inputs and Outputs}
    
    All the inputs and outputs for this model are continuous, we had a total of 27 inputs.
    \begin{itemize}
        \item X and Z positions of all the robots (Unity uses a y-up coordinate system) 
        \item Time in seconds since the start of the game 
        \item X and Z positions of the 10 closest rings, sorted in ascending order by distance
    \end{itemize}

    In order to retrieve the 10 closest rings, we applied the aforementioned frustum and occlusion culling on all the rings on the field. We then sorted them by distance and chose the first 10. All the other inputs are trivially obtained through Unity's GameObject API.
    
    The robots had memory of the last 10 inputs along with the current inputs. In Unity, this is referred to as "stacking vectors" and allows the model to make observations about inputs over a period of time. We decided to use this method over Recurrent Neural Nets-- which are an option in Unity-- to save on training time.
    
    
    The robot movement was implemented using simple velocity and rotation speed. These were the only 2 outputs of the model. This matches the control scheme that the robot drivers use in the competition.
    
    \subsection{Optimizations and Pre-processing}
    
    The utility of the localization algorithm is to have symmetrical inputs for robots on both alliances so that the input spaces are consistent between the red and blue teams in the VEX Competition. We found that, without these techniques, the model would tend to only converge for one alliance, due to differences in the input spaces between the alliances.

    We had a few requirements for the model inputs– it needed to know the position of the robot, and it needed to know the positions of as many as 10 rings visible to it, the input format should be simple to replicate on the real-world side, the inputs needed to be the same for each alliance, and the training must account for real-world noise.
    
    With this in mind, we decided on the following input post-processing on the robot's digital twin in the simulation.
    \begin{itemize}
        \item Robot Position
        \begin{itemize}
            \item Rotate it to the reference frame of the alliance (e.g., by 180 degrees if blue alliance, don’t rotate if red alliance) 
            \item Normalize values between –1 and 1 by dividing by half the field width 
            \item Add 10\% uniform noise during training to account for real-world noise, and for regularization
        \end{itemize}
        \item Ring Position
        \begin{itemize}
            \item Transform the position vector of each ring to the robot’s space using the inverse transform matrix of the robot space
            \item Remove the upward-facing axis 
            \item Divide by half field width 
            \item Add 10\% uniform noise 
        \end{itemize}
    \end{itemize}
    
    And on the real robot.
    \begin{itemize}
        \item Robot Position
        \begin{itemize}
            \item Rotate to alliance reference frame
            \item Divide by field width
        \end{itemize}
        \item Ring Position (detailed in 3.3)
        \begin{itemize}
            \item Add an axis to the pixel position and set it to 1 
            \item Multiply by the depth value 
            \item Inverse transform by the camera matrix (determined using OpenCV) which gives camera space coordinates 
            \item Transform the vector using a quaternion of the camera’s rotation to convert it to robot-space 
        \end{itemize}
    \end{itemize}
    
    \subsection{TensorBoard Logging}
    
    To ensure our simulation was trained properly and to keep track of our reward system we used TensorBoard to log our policy, reward, environment, and loss data. TensorBoard automatically generates graphs to monitor training.
    
    \begin{figure}[H]
        \centering
        \includegraphics[width=.8\textwidth]{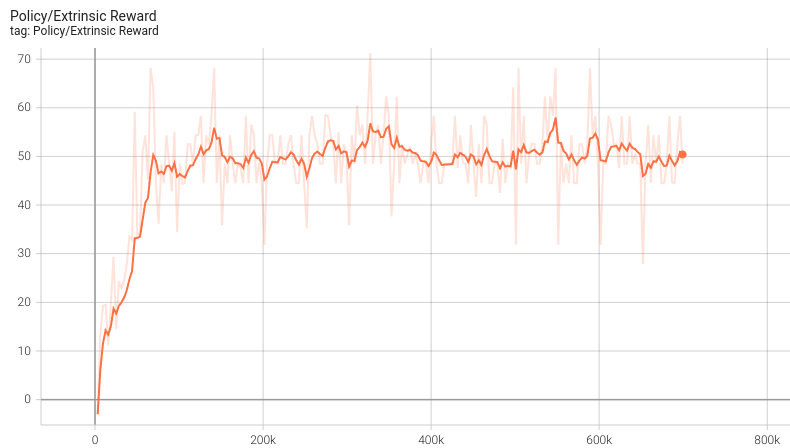}
        \caption{Mean episode reward over training steps}
        \label{fig:total_reward}
    \end{figure}

	\section{Object Tracking Methodology}
	\label{sec:cv}
	
	The Computer Vision team focused on the detection and localization of game objects \cite{rings_localization_mapping}. Our goal was to process this data to put it into our Reinforcement Learning model. We used an Intel Realsense D435 as our sole sensor for this. The Realsense camera gives us a color image as well as a depth map of how far each pixel is from the camera.

    It allows us to take an image, detect the game objects in the image, and calculate where those game objects are with relative ease. The detection of game objects is done using the color image, and the localization is done with the depth map. The culmination of our work can be found in the competition code \cite{competition_code} under the models folder. Developing the algorithms used in the competition was done in a separate repository \cite{rings_localization_mapping}. A video of the ring detection algorithm detailed in Section 3.2 can be found \underline{\href{https://www.youtube.com/watch?v=iHXsPqX9tcA&feature=youtu.be}{at this link}}. In this section we will be breaking down how the detection and localization algorithms work.  
    
    \subsection{YOLOv5s Ring Detector}
    
    We use a YOLOv5s model \cite{glenn_jocher_2020_4154370} trained on a custom ring dataset for the detection of rings. YOLOv5s was chosen due to its small size and familiarity among team members. The dataset to train the model was created from images taken on a phone of the rings in different positions, as well as screenshots from various competitions available on YouTube. We used Labelimg \cite{Labelimg} to label every ring in the photos. We made sure to capture images including robots and mobile goals to prevent over-fitting on circular features.  
    
    \begin{figure}[H]
        \centering
        \includegraphics[width=.45\textwidth]{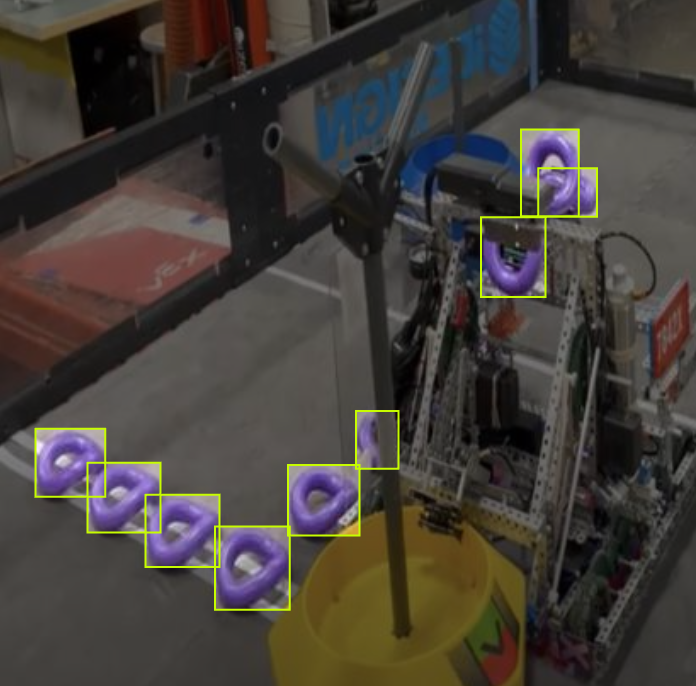}
        \hfill
        \includegraphics[width=.45\textwidth]{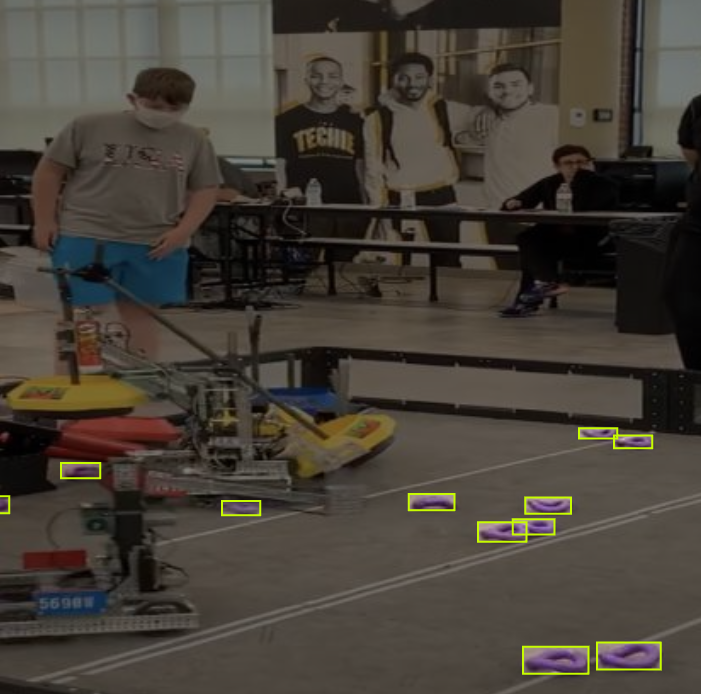}
        \caption{Annotated images using Labelimg}
        \label{fig:my_label}
    \end{figure}
    
    We then trained the model using Roboflow’s tutorial in a Google Colab notebook. Auto orient, resize, hue, blur, and noise data augmentations were added to improve accuracy in the real world. We were able to get decent results on our testing set. The model had high precision and recall.
    
    \begin{figure}[H]
        \centering
        \includegraphics[width=1\textwidth]{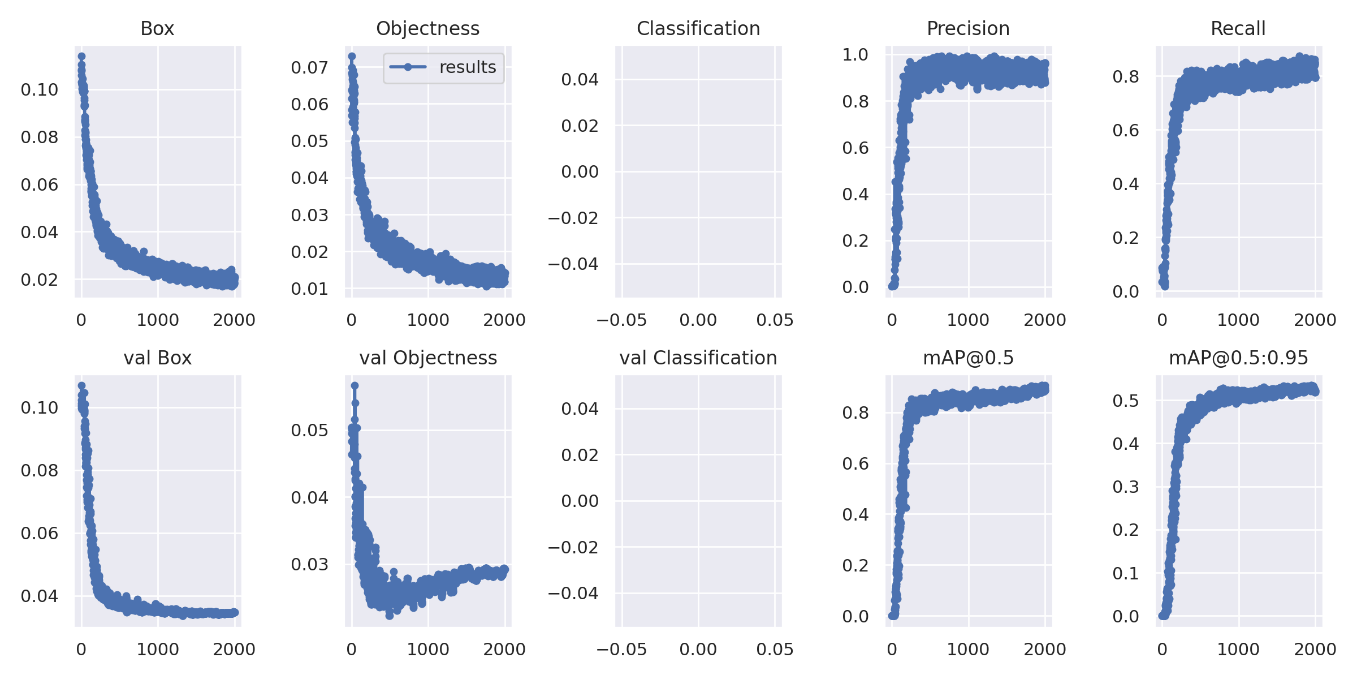}
        \caption{YOLOv5s Training statistics}
        \label{fig:my_label}
    \end{figure}
    
    Our fully trained model can be found in the competition code repository \cite{competition_code} in \textbf{\_models/best.pt}. With the Jetson Nano’s GPU, this model was able to run at about 10 inferences a second. In practice the model had a lot of false positives, to the point of it being unusable. This encouraged us to develop some pre-processing methods to detect “ring candidates” before passing it into the YOLOv5s model. Note that we could have also explored different methods, which are discussed in 5.2. Only looking at the sections of each image that has a “ring candidate” reduces the number of false positives we get. We used GRIP for the pre-processing.

    \subsection{YOLOv5s Ring Detector with GRIP}
    
    GRIP is a Graphically Represented Image Processing engine \cite{ClarkLeitschuh2016} created by Worcester Polytechnic Institute. It’s a tool used to rapidly visualize and deploy computer vision algorithms, and it’s commonly used by FIRST Robotics teams to create various vision systems. GRIP allows you to visualize how applying different algorithms will look to an image, and it will generate code for you to use. We use a combination of different techniques to select ring candidates.  

    The idea behind these preprocessing steps was to only look at pixels around purple objects in order to reduce the number of false positives we get in the real world. The processing is done in four steps. The parameters were obtained through trial-and-error.
    \newline
    
    Step 1: Capture an image using the Realsense camera. 
    
    \begin{figure}[H]
        \centering
        \includegraphics[width=.75\textwidth]{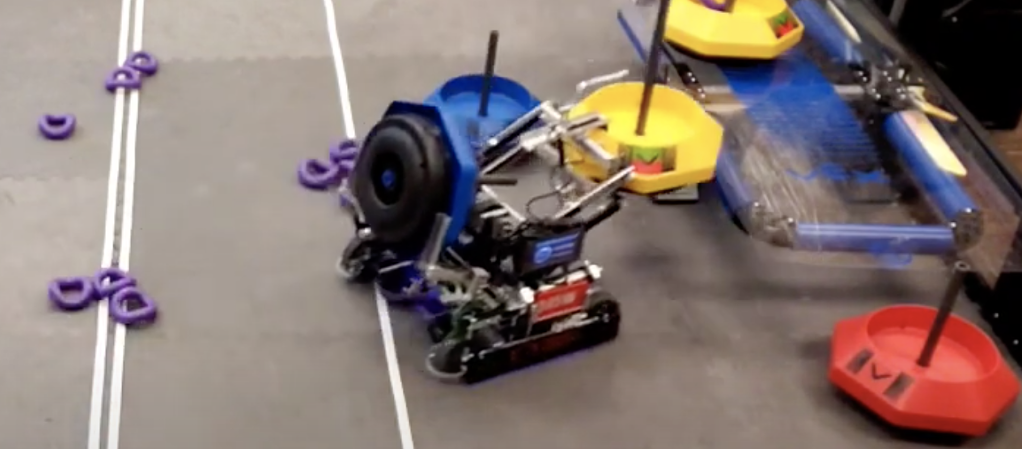}
        \caption{Image from RealSense camera}
        \label{fig:my_label}
    \end{figure}
    
    Step 2: Apply a HSV threshold to the original image. Hue: 123-169. Saturation: 39-192. Value: 76-255.

    \begin{figure}[H]
        \centering
        \includegraphics[width=.75\textwidth]{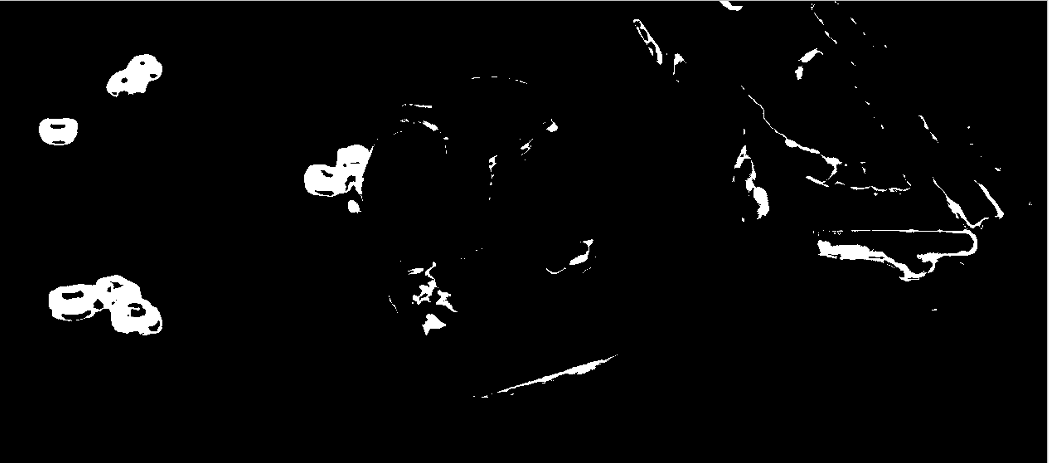}
        \caption{Figure 6 after HSV threshold applied}
        \label{fig:my_label}
    \end{figure}
    
    Step 3: Apply a blur to the HSV threshold. Box blur. Radius: 17. 
    
    \begin{figure}[H]
        \centering
        \includegraphics[width=.75\textwidth]{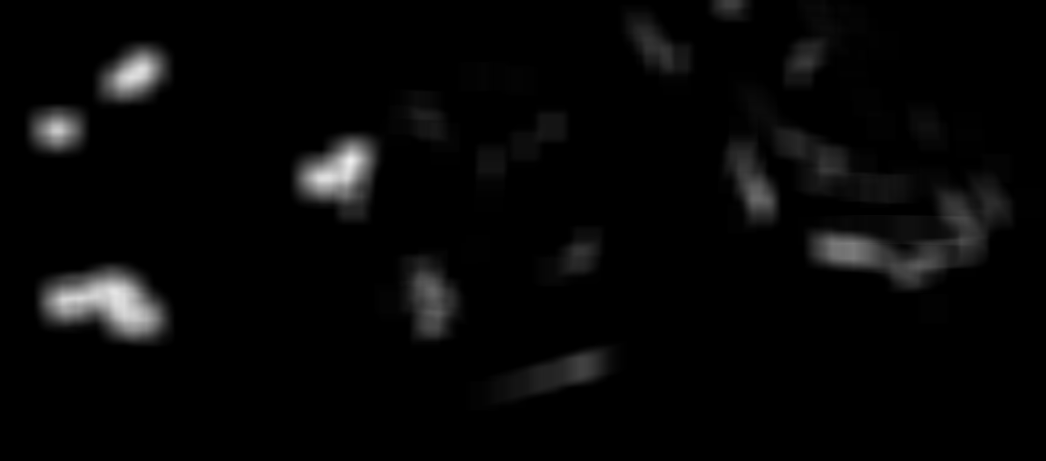}
        \caption{Figure 7 after Blur applied}
        \label{fig:my_label}
    \end{figure}
    
    Step 4: Mask the original input with the blurred HSV threshold. 
    
    \begin{figure}[H]
        \centering
        \includegraphics[width=.75\textwidth]{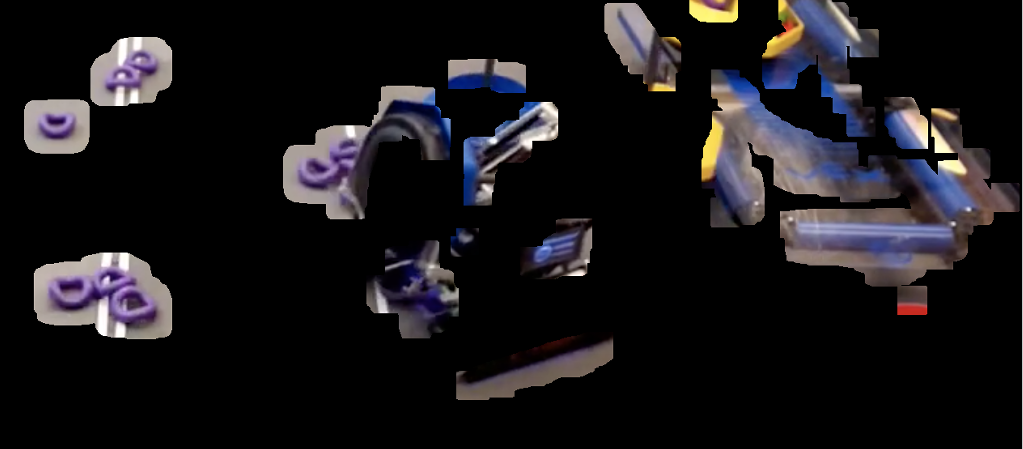}
        \caption{Figure 6 masked with Figure 8}
        \label{fig:my_label}
    \end{figure}
    
    This greatly improved our performance in the real world, and we found this was the best way to consistently detect rings, since it effectively omits anything in the background. A video of the ring detector in use can be found here.

    The pipeline set up in GRIP looks like this.
    
    \begin{figure}[H]
        \centering
        \includegraphics[width=.75\textwidth]{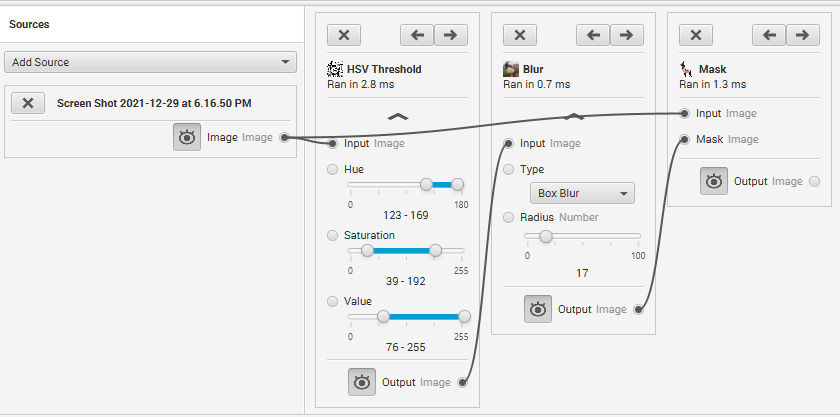}
        \caption{GRIP Pipeline used}
        \label{fig:my_label}
    \end{figure}
    
    \subsection{Ring Localization}
    
    This is the final step in the Computer Vision pipeline. Once we have a ring’s position in the Realsense’s image, we then have to find the ring’s relative position to the robot to match the training environment in Section 2.3.
    
    
    For each ring we detect, we have its position in the camera’s photo and how far away it is from the camera using the Realsense’s depth map.
    
    
    
    \begin{figure}[H]
        \centering
        \includegraphics[width=.75\textwidth]{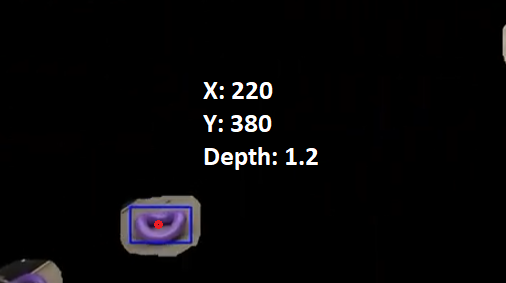}
        \caption{Example ring localization data}
        \label{fig:my_label}
    \end{figure}
    
    The camera matrix $K$ is obtained through OpenCV's camera calibration functionality. We printed out a reference checkerboard image, and took pictures of it with the RealSense camera. We then loaded the images in a Python script and called the OpenCV function which emitted a camera matrix in pixel units. 
    
    From here we can get the ring’s position in the camera space by multiplying the ring’s position in the image by the depth.
    
    $$v = \begin{bmatrix} 220 & 380 & 1 \end{bmatrix} \ast 1.2$$

    We then dot product the inverse camera matrix, $K^{-1}$, with the ring’s position in the camera space, $v$. This will give us the coordinates of the ring in real space. 
    
    $$r = K^{-1} \cdot v$$
    
    Now that we have the ring’s position in real space, we’ll have to perform another rotation about the x-axis to account for the camera’s tilt. We will call the space relative to the robot the world space. 
    
    $$w = \text{rotate(}r\text{, -tilt, x-axis)}$$
    
    We can then use the x and z coordinates of $w$ as the real position of each ring with respect to the robot. This localization algorithm is repeated for every ring detected. We then sort the rings by distance for the Reinforcement Learning model input.

	\section{Brain to Jetson Nano Data Pipeline Methodology}
	\label{sec:pipeline}
	
	The hardware pipeline team focused on developing a protocol to transfer data between a VEX V5 Brain and a NVIDIA Jetson Nano \cite{competition_code}. Offloading the data for processing is necessary due to the weak processing power of the VEX V5 Brain. The NVIDIA Jetson Nano has a GPU which allows the AI Pipeline-- which predicts ring positions and robot movement-- to quickly perform its calculations. We used Purdue University's PROS Robotics Operating System \cite{PROS} to aid in data transfer.
	
	
	\begin{figure}[H]
	    \centering
	    \includegraphics[width=.5\textwidth]{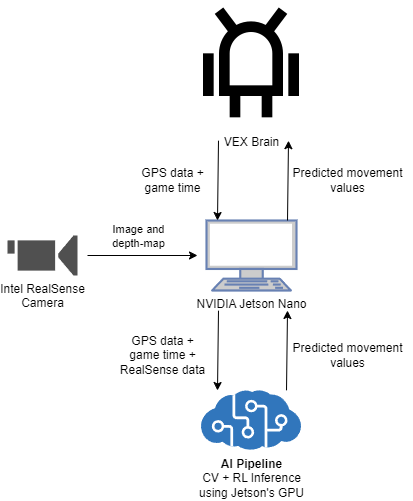}
	    \caption{Brain to Jetson Nano Data Pipeline System diagram}
	    \label{fig:my_label}
	\end{figure}
	
	\subsection{Setup}
	
	We developed scripts set up to assist in starting the data transfer pipeline. We use the PROS Terminal to print to stdout on the Brain side of the protocol and receive data over USB on the Jetson side.  

    One issue we ran into was starting the PROS Terminal alongside our main Python script, which ran the computer vision pipeline and reinforcement learning model inference. They both need their own instances of a shell. To overcome this problem, we used TMUX. TMUX (Terminal multiplexer) allows us to run multiple instances of the shell and therefore run both the protocol and the PROS Terminal simultaneously. 
    
    We can use two different methods to start the data transfer. One way is to SSH into the Jetson and manually run the startup script. The other method is setting up a startup cronjob that will run the startup script once the Jetson boots. The second method is preferred since internet access is not always guaranteed. 
    
    \subsection{Packets}
    
    The packets are in the form of a space separated string encoding the data, and an additional iterator at the end. To mitigate the transfer of identical packets the protocol checks to make sure that the packet it reads is different than the last one processed. This system is not ideal, and thoughts on how to improve it are listed in Section 5.3.
    
    \subsection{VEX V5 Brain Writing}
    
    The protocol begins with the Brain sending data. For the AI pipeline to be able to predict the robot's future movement it needs the location of the robot and the game time. There is a program running on the Brain which gathers this data from the robot's game positioning system sensor-- made by VEX-- and the game clock. This data is sent to the protocol where it is packed into a packet and sent off to the Jetson. The packets are printed to stdout where the PROS terminal can send them to the Jetson. During this stage the Jetson is constantly awaiting a message from the Brain, it will stay in this mode until it receives a message. 
    
    \subsection{NVIDIA Jetson Nano Reading and Writing}
    
    The Jetson receives the packet from the Brain. Once the packet is unpacked the data is sent to the AI pipeline. The AI pipeline predicts what the velocity and rotation of the robot should be. These values are packaged up into a packet by the protocol and sent back off to the Brain. During this stage the Brain is awaiting a message from the Jetson and will stay in this mode until it receives a message.

    During this step the protocol takes advantage of the PROS terminal. An instance of this is ran on the Jetson which reads incoming packets from stdout sent by the Brain. These packets are saved to a file where they can later be read. 
    
    \subsection{VEX V5 Brain Reading}
    
    Once the Brain receives the packet it unpacks the velocity and rotation provided by the AI pipeline. The velocity and rotation are used to adjust the drivetrain’s motor speeds. The protocol repeats from Section 4.3 until the game is finished. 
	
	\section{Discussion and Results}
	\label{sec:discussion}
	
	\subsection{Simulation}
	
	Our team understood that a Simulation-to-Real training pipeline was unnecessary for this problem. Once rings are detected and localized, one could use an algorithm like A* \cite{ASTAR} for motion planning to collect them all. The team had hopes that a reinforcement learning model in this setting would learn more complicated behaviors, we anticipated defense strategies to emerge once agents maxed out the amount of rings they could collect. This, unfortunately, was not the case. Additional reward function and simulation engineering might be necessary to see behaviors like these emerge. These behaviors could also emerge with a longer training time, or a less complex action space.
	
	\begin{figure}[H]
        \centering
        \includegraphics[width=.8\textwidth]{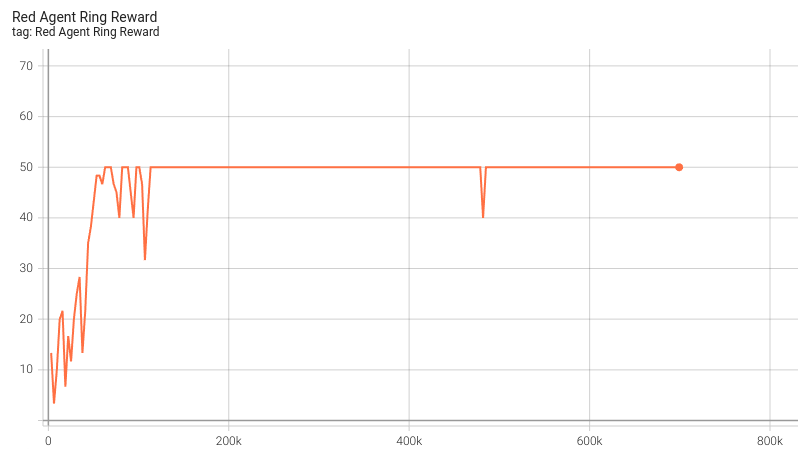}
        \caption{Mean ring reward over training steps}
        \label{fig:mean_ring_reward}
    \end{figure}
	
	Despite this, we observed our agent collect rings in both the simulation and real world well. Our agent begins to consistently max out the number of rings it can hold after only 113.4k steps of training (90 games), as shown in Figure 13. We did not observe a considerable performance increase after this point. Rewards 2, 3, and 4-- detailed in Figure \ref{fig:reward_fcn}-- seemed to have minimal effect on the agent's behavior. The mean episode reward slightly deviated around the max ring reward until the end of training, shown in Figure \ref{fig:total_reward}.
	
	\begin{figure}[H]
    \begin{center}
    \begin{tabular}{ |c|c|c|c| } 
    \hline
    Reward & Value & Range \\
    \hline
    (1) Rings collected by the agent, max 10 rings & 5 per ring & [0, 50] \\ 
    \hline
    (2) Pinning the opposing agent for 5 seconds, disqualifying &-5 per pin & [-5, 0] \\the agent and restarting the episode & &  \\ 
    \hline
    (3) Mobile goal scored in the agent's side of the field & 12 per goal & [0, 36] \\at the end of the episode, 2 alliance 1 neutral  && \\ 
    \hline
    (4) Position penalty for ending on the opposing alliance's side & -17.5 & [-17.5, 0]\\
    \hline
    \end{tabular}
    \end{center}
    \caption{Reward function}
    \label{fig:reward_fcn}
    \end{figure}

	
	
    
    We recommend teams use a more traditional approach to planning and control if they plan to make a competitive robot. 
    
    \subsection{Object Tracking}
	
	The computer vision pipeline performed well in testing as well as in live performance. The pipeline we developed could be repurposed for a multitude of different things in other competitions and exists as a standalone part of the system anyone could modify with relative ease. The system is not perfect, however. In the future we would recommend teams:
	\begin{itemize}
	    \item Gather more data and get diverse photos. 
	    \item Experiment with pre-trained weights. 
	    \item Experiment with classical computer vision methods. 
	    \item Experiment with vanilla CNNs or smaller models. 
	    \begin{itemize}
	        \item Our model likely had too many parameters for the problem. Detecting a circular object like a ring is not very difficult to learn. If you increase bias by using a less complicated model, you’ll likely reduce how much the model overfits. 
	        \item Previous work had much better results using only a CNN with a larger input size and more labeled data \cite{TaborGilgenbachBerry2018}. 
	    \end{itemize}
	    \item Experiment with models which might better suited for small object detection, such as iSmallNet \cite{ismallnet} or YOLOX \cite{YOLOX}. 
	    \item Increase the input size of the model if you’re trying to detect smaller objects. 
	    \item Turn off the mosaic data augmentation. 
	    \begin{itemize}
	        \item This is usually done for general object detectors to learn how to identify parts of an object, it likely hurt our performance while training. 
	    \end{itemize}
	\end{itemize}
	
	\subsection{Brain to Jetson Pipeline}
	


    
    
    
    The pipeline worked well as long as it was able to start. In reality-- it can take a long time before the robot starts moving if the startup cronjob is used. This is due to the Jetson start up time. If teams choose to use this method, we recommend turning on the V5 Brain and Jetson Nano least a minute before your game. 

    A better startup process with or without the cronjob is room for future development within this system and PROS. Creating a linux init.d script might be more convenient than a cronjob. We are also currently relying on TMUX to run multiple terminal windows due to incompatibility with PROS Terminal.

	
	
	
	
	
	
	\printbibliography

	\appendix
	
\end{document}